\documentclass[graybox]{svmult}

\usepackage{mathptmx}       
\usepackage{helvet}         
\usepackage{courier}        
\usepackage{type1cm}        
\usepackage{makeidx}         
\usepackage{graphicx}        
\usepackage{multicol}        
\usepackage[bottom]{footmisc}

\usepackage{times}
\usepackage{epsfig}
\usepackage{graphicx}
\usepackage{amsmath}
\usepackage{amssymb}
\usepackage{fancyhdr}
\usepackage[utf8]{inputenc}
\usepackage{setspace}
\usepackage{algpseudocode}
\usepackage{algorithm}
\usepackage{algorithmicx}
\usepackage{mathtools}
\usepackage{boxedminipage}
\usepackage{graphicx}
\usepackage{verbatim}
\usepackage{dsfont}
\usepackage{multirow}
\usepackage{amsfonts}
\usepackage[T1]{fontenc}
\usepackage{url}
\usepackage{epsfig}
\usepackage{epstopdf}
\usepackage{caption}
\usepackage{adjustbox}
\usepackage{booktabs}

\makeindex             

\begin{document}
	
	\title*{Deep Learning Architectures for Face Recognition in Video Surveillance}
	\author{Saman Bashbaghi, Eric Granger, Robert Sabourin and Mostafa Parchami}
	\authorrunning{Saman Bashbaghi et al.}
	\institute{Saman Bashbaghi, Eric Granger, Robert Sabourin \at Laboratoire. d'imagerie de vision et d'intelligence artificielle,\\
		\'{E}cole de technologie sup\'{e}rieure, Universit\'{e} du Qu\'{e}bec, Montreal, Canada\\ \email{bashbaghi@livia.etsmtl.ca, {eric.granger, robert.sabourin}@etsmtl.ca}
		\and Mostafa Parchami \at Computer Science and Engineering Department, University of Texas at Arlington, TX, USA\\ \email{mostafa.parchami@mavs.uta.edu}}
	%
	%
	\maketitle
	
	\abstract*{Abstract}
	
	\abstract{Face recognition (FR) systems for video surveillance (VS) applications attempt to accurately detect the presence of target individuals over a distributed network of cameras. In video-based FR systems, facial models of target individuals are designed a priori during enrollment using a limited number of reference still images or video data. These facial models are not typically representative of faces being observed during operations due to large variations in illumination, pose, scale, occlusion, blur, and to camera inter-operability. Specifically, in still-to-video FR application, a single high-quality reference still image captured with still camera under controlled conditions is employed to generate a facial model to be matched later against lower-quality faces captured with video cameras under uncontrolled conditions. Current video-based FR systems can perform well on controlled scenarios, while their performance is not satisfactory in uncontrolled scenarios mainly because of the differences between the source (enrollment) and the target (operational) domains. Most of the efforts in this area have been toward the design of robust video-based FR systems in unconstrained surveillance environments. This chapter presents an overview of recent advances in still-to-video FR scenario through deep convolutional neural networks (CNNs). In particular, deep learning architectures proposed in the literature based on triplet-loss function (e.g., cross-correlation matching CNN, trunk-branch ensemble CNN and HaarNet) and supervised autoencoders (e.g., canonical face representation CNN) are reviewed and compared in terms of accuracy and computational complexity.}
	
	\section{Introduction}
	\label{sec:Intro}
	
	Face recognition (FR) systems in video surveillance (VS) has received a significant attention during the past few years. Due to the fact that the number of surveillance cameras installed in public places is increasing, it is important to build robust video-based FR systems \cite{Zheng2017}. In VS, capture conditions typically range from semi-controlled with one person in the scene (e.g. passport inspection lanes and portals at airports), to uncontrolled free-flow in cluttered scenes (e.g. airport baggage claim areas, and subway stations). Two common types of applications in VS are: (1) still-to-video FR (e.g., watch-list screening), and (2) video-to-video FR (e.g., face re-identification or search and retrieval) \cite{Bashbaghi2017, DelaTorre2014, pagano2014adaptive}. In the former application, reference face images or stills of target individuals of interest are used to design facial models, while in the latter, facial models are designed using faces captured in reference videos. This chapter is mainly focused on still-to-video FR with a single sample per person (SSPP) under semi- and unconstrained VS environments.
	
	The number of target references is one or very few in still-to-video FR applications, and the characteristics of the still camera(s) used for design significantly differ from the video cameras used during operations \cite{BashbaghiPR}. Thus, there are significant differences between the appearances of still ROI(s) and ROIs captured with surveillance cameras, according to various changes in ambient lighting, pose, blur, and occlusion \cite{barr2012face, Matta2009180}. During enrollment of target individuals, facial regions of interests (ROIs) isolated in reference still images are used to design facial models, while during operations, the ROIs of faces captured in videos are matched against these facial models. In VS, a person in a scene may be tracked along several frames, and matching scores may be accumulated over a facial trajectory (a group of ROIs that correspond to the same high-quality track of an individual) for robust spatio-temporal FR \cite{Dewan2016}.
	
	In general, methods proposed in the literature for still-to-video FR can be broadly categorized into two main streams: (1) conventional, and (2) deep learning methods. The conventional methods rely on hand-crafted feature extraction techniques and a pre-trained classifier along with fusion, while deep learning methods automatically learn features and classifiers cojointly using massive amounts of data. In spite of improvements achieved using the conventional methods, yet they are less robust to real-world still-to-video FR scenario. On the other hand, there exists no feature extraction technique that can overcome all the challenges encountered in VS individually \cite{Bashbaghi2017, Huang2015, Taigman_2014_CVPR}.
	
	Conventional methods proposed for still-to-video FR are typically modeled as individual-specific face detectors using one- or 2-class classifiers in order to enable the system to add or remove other individuals and easily adapt over time \cite{bashbaghiwatch, pagano2014adaptive}. Modular systems designed using individual-specific ensembles have been successfully applied in VS \cite{DelaTorre2014, pagano2014adaptive}. Thus, ensemble-based methods have been shown as a reliable solution to deal with imbalanced data, where multiple face representations can be encoded into ensembles of classifiers to improve the robustness of still-to-video FR \cite{Bashbaghi2017}. Although it is challenging to design robust facial models using a single training sample, several approaches have addressed this problem, such as multiple face representations, synthetic generation of virtual faces, and using auxiliary data from other people to enlarge the training set \cite{bashbaghiwatch, Kamgar2011, kan2013adaptive, SVDL}. These techniques seek to enhance the robustness of face models to intra-class variations. In multiple representations, different patches and face descriptors are employed \cite{bashbaghiwatch, Bashbaghi2017}, while 2D morphing or 3D reconstructions are used to synthesize artificial face images \cite{Kamgar2011, Mokhayeri2015}. A generic auxiliary dataset containing faces of other persons can be exploited to perform domain adaptation \cite{Ma2015}, and sparse representation classification through dictionary learning \cite{SVDL}. However, techniques based on synthetic face generation and auxiliary data are more complex and computationally costly for real-time applications, because of the prior knowledge required to locate the facial components reliably, and the large differences between the quality of still and video ROIs, respectively.
	
	Recently, several deep learning based solutions have been proposed to learn effective face representations directly from training data through convolutional neural networks (CNNs) and nonlinear feature mappings \cite{Chellappa2016, Huang2012, Schroff_2015_CVPR, Sun_2013_ICCV, Sun_2014_CVPR}. In such methods, different loss functions can be considered in the training process to enhance the inter-personal variations, and simultaneously reduce the intra-personal variations. They can learn non-linear and discriminative feature representations to cover the existing gaps compared to the human visual system \cite{Taigman_2014_CVPR}, while they are computationally costly and typically require a large number of labeled data to train. To address the SSPP problem in FR, a triplet-based loss function have been introduced in \cite{Ding2016, Parchami2017CCM, Parchami2017, parkhi2015deep, Schroff_2015_CVPR} to discriminate between a pair of matching ROIs and a pair of non-matching ROIs. Ensemble of CNNs, such as trunk-branch ensemble CNN (TBE-CNN) \cite{Ding2016} and HaarNet \cite{Parchami2017} have been shown to extracts features from the global appearance of faces (holistic representation), as well as, to embed asymmetrical features (local facial feature-based representations) to handle partial occlusion. Moreover, supervised autoencoders have been proposed to enforce faces with variations to be mapped to the canonical face (a well-illuminated frontal face with neutral expression) of the person in the SSPP scenario to generate robust feature representations \cite{Gao2015, Parchami2017CFR}.
	
	\section{Bachground of Video-Based FR Through Deep Learning}
	\label{sec:Background}
	
	Deep CNNs have recently demonstrated a great achievement in many computer vision tasks, such as object detection, object recognition, etc. Such deep CNN models have shown to appropriately characterize different variations within a large amount of data and to learn a discriminative non-linear feature representation. Furthermore, they can be easily generalized to other vision tasks by adopting and fine-tuning pre-trained models through transfer learning \cite{Chellappa2016, Schroff_2015_CVPR}. Thus, They provide a successful tool for different applications of FR by learning effective feature representations directly from the face images \cite{Chellappa2016, Huang2012, Schroff_2015_CVPR}. For example, DeepID, DeepID2, and DeepID2+ have been proposed in \cite{Sun_NIPS2014_5416, Sun_2014_CVPR, Sun_2015_CVPR}, respectively, to learn a set of discriminative high-level feature representations.
	
	For instance, an ensemble of CNN models was trained in \cite{Sun_2014_CVPR} using the holistic face image along with several overlapping/non-overlapping face patches to handle the pose and partial occlusion variations. Fusion of these models is typically carried out by feature concatenation to construct over-complete and compact representations. Followed by \cite{Sun_2014_CVPR}, feature dimension of the last hidden layer was increased in \cite{Sun_NIPS2014_5416, Sun_2015_CVPR}, as well as, exploiting supervision to the convolutional layers in order to learn hierarchical and non-linear feature representations. These representations aim to enhance the inter-personal variations due to extraction of features from different identities separately, and simultaneously reduce the intra-personal variations. In contrast to DeepID series, an accurate face alignment was incorporated in Microsoft DeepFace \cite{Taigman_2014_CVPR} to derive a robust face representation through a nine-layer deep CNN. In \cite{Sun_2013_ICCV}, the high-level face similarity features were extracted jointly from a pair of faces instead of a single face through multiple deep CNNs for face verification applications. Since these approaches are not considered variations like blurriness and scale changes (distance of the person from surveillance cameras), they are not fully adapted for video-based FR applications.
	
	Similarly, for the SSPP problems, a triplet-based loss function has been lately exploited in \cite{Ding2016, Parchami2017CCM, Parchami2017, parkhi2015deep, Schroff_2015_CVPR} to learn robust face embeddings, where this type of loss seeks to discriminate between the positive pair of matching facial ROIs from the negative non-matching facial ROI. A robust facial representation learned through triplet-loss optimization has been proposed in \cite{Parchami2017CCM} using a compact and fast cross-correlation matching CNN (CCM-CNN). However, CNN models like the trunk-branch ensemble CNN (TBE-CNN) \cite{Ding2016} and HaarNet \cite{Parchami2017} can further improve robustness to variations in facial appearance by the cost of increasing computational complexity. In such models, the trunk network extracts features from the global appearance of faces (holistic representation), while the branch networks embed asymmetrical and complex facial traits. For instance, HaarNet employs three branch networks based on Haar-like features, while facial landmarks are considered in TBE-CNN. However, these specialized CNNs represent complex solutions that are not perfectly suitable for real-time FR applications \cite{Canziani2016}.
	
	Moreover, autoencoder neural networks can be typically employed to extract deterministic non-linear feature mappings robust to face images contaminated by different noises, such as illumination, expression and poses \cite{Gao2015, Parchami2017CFR}. An autoencoder network contains encoder and decoder modules, where the former module embed the input data to the hidden nodes, while the latter returns the hidden nodes to the original input data space with minimizing the reconstruction error(s) \cite{Gao2015}. Several autoencoder networks inspired from \cite{vincent2010} have been proposed to remove the aforementioned variances in face images \cite{Gao2015, Kan2014, Le2013}. These networks deal with faces containing different types of variations (e.g., illumination, pose, etc.) as noisy images. For instance, a facial component-based CNN has been learned in \cite{zhu2014recover} to transform faces with changes in pose and illumination to frontal view faces, where pose-invariant features of the last hidden layer are employed as face representations. Similarly, several deep architecture have been proposed using multi-task learning in order to rotate faces with arbitrary poses and illuminations to target-pose faces \cite{Yim2015, Zhu2014NIPS}. In addition, a general deep architecture was introduced in \cite{Ghodrati2016} to encode a desired attribute and combine it with the input image to generate target images as similar as the input image with a visual attribute (a different illumination, facial appearance or new pose) without changing other aspects of a face.
	
	\section{Deep Learning Architectures for FR in VS}
	\label{sec:Technology}
	
	In this section, the most recent deep learning architectures proposed for video-based FR considering the SSPP problem are addressed. These architectures can be categorized into two groups: (1) Deep CNN models trained using triplet-loss function, and (2) deep autoencoders.
	
	\subsection{Deep CNNs Using Triplet-Loss}
	
	Recently, deep learning algorithms specialized for FR mostly utilize triplet-loss in order to train the deep architecture and thereby learning a discriminant face representation \cite{Ding2016, Schroff_2015_CVPR}. However, careful triplet sampling is a crucial step to achieve a faster convergence \cite{Schroff_2015_CVPR}. In addition, employing triplet-loss is challenging since the global distributions of the training samples are neglected in optimization process.
	
	Triplet-loss approach was first proposed in \cite{Schroff_2015_CVPR} to train CNNs for robust face verification. To that end, the representation of triplets (three faces containing an anchor and a positive image of the same subject and a negative image of other subjects) are $L_2$-normalized as the input of triplet-loss function. It therefore ensures that the input representations of face images lie on a unit hypersphere prior to apply triplet-loss function \cite{Ding2016}. Deep CNN models proposed for video-based FR that employed triplet-loss for training are reviewed in the following subsections.
	
	\subsubsection{Cross-Correlation Matching CNN}
	
	An efficient deep CNN architecture has been proposed in \cite{Parchami2017CCM} for still-to-video FR from a single reference facial ROI per target individual. Based on a pair-wise cross-correlation matching (CCM) along with a robust facial representation learned through triplet-loss optimization, CCM-CNN is a fast and compact network (requires few branches, layers and parameters). It exploits a matrix Hadamard product followed by a fully connected layer that simulates the adaptive weighted cross-correlation \cite{heo2011robust}. A triplet-based optimization approach has been employed to learn discriminant representations based on triplets containing the positive, negative video ROIs and the corresponding still ROI. In particular, the similarity between the representations of positive video ROIs and the reference still ROI is enhanced, while the similarity between negative video ROIs and the both reference still and positive video ROIs is increased. To further improve robustness of facial models, the CCM-CNN fine-tuning process incorporates a diverse knowledge by generating synthetic faces based on still and video ROIs of non-target individuals.
	
	\begin{figure}[h]
		\begin{center}
			\includegraphics[width=0.67\textwidth]{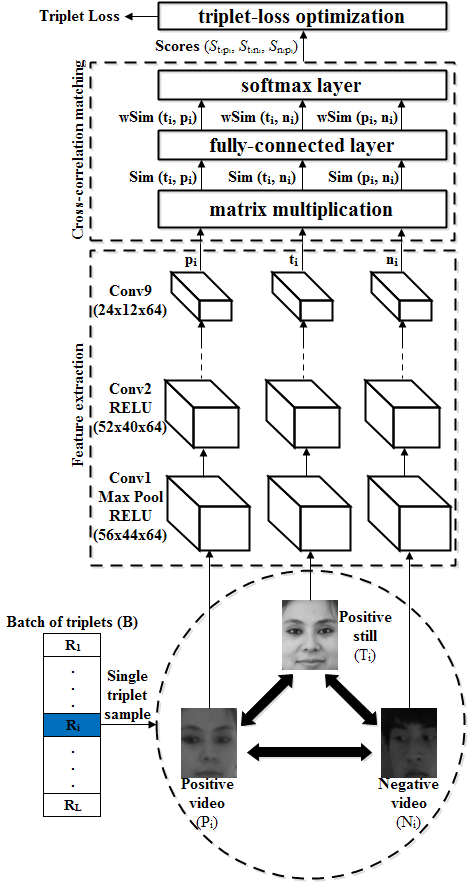}
		\end{center}
		\caption{Training pipeline of the CCM-CNN \cite{Parchami2017CCM}.}
		\label{fig:Training}
	\end{figure}
	
	As shown in Figure \ref{fig:Training}, the CCM-CNN learns a robust facial representation by iterating over a batch of training triplets $B = \left\{ R_1, \ldots, R_L \right\}$ $= \left\{ {\left( {{T_1},{P_1},{N_1}} \right), \ldots ,\left( {{T_L},{P_L},{N_L}} \right)} \right\}$, where $L$ is the batch size, and each triplet $R_i$ contains a still ROI $T_i$ along with a corresponding positive ROI $P_i$ and a negative ROI $N_i$ from operational videos. This architecture was inspired by Siamese networks containing identical subnetworks with the same configurations, parameters and weights. Therefore, fewer parameters are required for training that can avoid overfitting. The CCM-CNN consists of three main components -- feature extraction, cross-correlation matching and triplet-loss optimization. The feature extraction pipeline extracts discriminative feature maps from ROIs that are similar for two images of the same person under different capture conditions (e.g., illumination and pose). The cross-correlation matching component inputs feature maps extracted from the ROIs and calculates the likelihood of the faces belonging to the same person. Finally, triplet-loss optimization computes a loss function to maximize similarity of the still ROIs and their respective positive samples in the batch, while minimizing similarity between still ROIs and their negative ROIs, as well as, positive and negative ROIs.
	
	Despite differences in the domains between reference target still ROIs and target/non-target video ROIs, the CCM-CNN can effectively extract discriminant features. As shown in Figure \ref{fig:Training}, feature extraction is carried out by 3 identical subnetworks for still, positive and negative faces. These subnetworks process three input faces and the weights are shared across them. Each subnetwork consists of 9 convolutional layers each followed by a spatial batch normalization, drop-out, and RELU layers. Contrary to former convolutional layers, the last convolutional layer is not followed by a RELU in order to maintain the representativeness of the final feature map and to avoid losing informative data for the matching stage. Moreover, a single max-pooling layer is added after the first convolution layer to increase the robustness to small translation of faces in the ROI.
	
	In the CCM-CNN, all three feature extraction pipelines share the same set of parameters. This ensures that the features extracted from target still ($\mathbf{t_i}$), positive ($\mathbf{p_i}$) and negative ($\mathbf{n_i}$) are consistent and comparable. Each convolutional layer has 64 filters of size 5x5 without padding. Thus, given the input size of 120x96, the output of each branch is of size $N_f= \text{24x12x64}$ features.
	
	After extracting features from the still and video ROIs, a pixel-based matching method is employed to effectively compare these feature maps and measure the matching similarity. The process of comparison in the CCM-CNN has three stages: matrix Hadamard product, fully connected neural network, and finally a softmax. Instead of concatenating feature vectors of different branches as input to the fully connected layer, the feature maps representing the ROIs are multiplied with each other to encode pixel-wise correlation between each pair of ROI in the triplet. This approach eliminates the complexity of matching by replacing the concatenation with a simple element-wise matrix multiplication and directly encodes similarity as opposed to let the network learn how to match input concatenated feature vectors.
	
	The matrix Hadamard product is exploited to simulate cross-correlation, where Hadamard product of the two matrices provides a single feature map that represents the similarity of the two ROIs. For example, the similarity $Sim(\mathbf{t}_{i}, \mathbf{p}_{i})$ and cross-correlation $\mathbf{w}Sim$($\mathbf{t}_{i}$, $\mathbf{p}_{i}$) of still $\mathbf{t}_{i}$ and positive $\mathbf{p}_{i}$ feature maps is computed as follows, respectively, using matrix Hadamard product:
	\begin{equation}
	Sim (\mathbf{t}_{i}, \mathbf{p}_{i}) = (\mathbf{t}_{i} \odot \mathbf{p}_{j})
	\end{equation}
	\begin{equation}
	wSim (\mathbf{t}_{i}, \mathbf{p}_{i}) = \mathbf{\omega}_{m} \cdot RELU(\mathbf{\omega}_{n} \cdot Sim(\mathbf{t}_{i}, \mathbf{p}_{i})+ \mathbf{b}_{n}) + \mathbf{b}_{m}
	\end{equation}
	\hspace*{-0.17cm} where $\mathbf{\omega}_{m}$, $\mathbf{\omega}_{n}$, $\mathbf{b}_{m}$ and $\mathbf{b}_{n}$ are the weights and biases of the two fully-connected layers applied to the vectorized output of the matrix multiplication. Furthermore, a softmax layer is applied to obtain a probability-like similarity score for each of the two classes (match and non-match).
	
	A multi-stage approach is considered to efficiently train the CCM-CNN based on reference still ROI and operational videos. To that end, pre-training is performed using a large generic FR dataset, and a domain specific dataset for still-to-video FR is used for fine-tuning. To that end, a set of matching and non-matching images is selected from the Labeled Faces in the Wild (LFW)~\cite{LFWTech}. Images from this set are augmented to roughly 1.3M training triplets. In order to consistently update the set of training triplets, the on-line triplet sampling method \cite{Schroff_2015_CVPR} is used for 50 epochs.
	
\begin{figure}[h!]
		\begin{center}
			\includegraphics[width=0.67\textwidth]{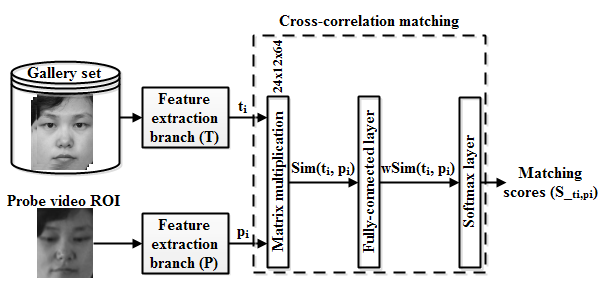}
		\end{center}
		\caption{The operational phase of the CCM-CNN \cite{Parchami2017CCM}.}
		\label{fig:operational_CCM}
	\end{figure}

	In contrast with FaceNet~\cite{Schroff_2015_CVPR}, a pair-wise triplet-loss optimization function was proposed to effectively train the network. In order to adapt the network for pairwise triplet-based optimization, it is modified by incorporating additional feature extraction branches. Each batch contains several triplets, and for each triplet, the network seeks to learn the correct classification. During the training, each branch of the feature extraction pipeline is assigned to a component of the triplet -- the main branch is responsible for processing the reference still ROI, while the positive (negative) branch extracts features from the positive (negative) video ROI of the triplet. Moreover, the cross correlation matching pipeline is modified to benefit from the triplets by introducing an Euclidean loss layer followed by softmax which computes the similarity for each pair of ROIs in the triplet. The loss layer is exploited to compute the overall loss of the network as follows:
	\begin{equation}
	\mbox{\small{Triplet Loss}} = \small{\frac{1}{L} \sum_{R_i \in B} \sqrt {{{\left( {1 - {S_{t_{i}p_{i}}}} \right)}^2} + S_{t_{i}n_{i}}^2 + S_{n_{i}p_{i}}^2}}
	\end{equation}
	\hspace*{-0.17cm} where $S_{tp}$, $S_{tn}$, and $S_{np}$ are the similarity scores from cross-correlation matching between (1) the reference (positive) still ROI and positive video ROI, (2) still ROI and negative video ROI, and (3) negative and positive video ROIs of the triplet, respectively, computed using the aforementioned approach. During operations (see Figure \ref{fig:operational_CCM}) the additional feature extraction branch (negative branch, N) is removed from the network, and only the still and the positive branches (P) are taken into account. Thus, the main branch (T) extracts features from a  reference still ROIs, while the positive branch extracts features from the probe video ROI to determine whether they belong to the same person.
	
	During fine-tuning, CCM-CNN acquires knowledge on the similarities and dissimilarities between the target individuals of interest enrolled to the system. In order to improve the robustness of facial models intra-class variation, the network is fine-tuned with synthetic facial ROIs generated from the high-quality still ROIs that account for the operation domain. For each still image, a set of augmented images are generated using different transformations, such as shearing, mirroring, rotating and translating the original still image. In contrast with the pre-training, the focus of the fine-tuning stage is to learn dissimilarities between the subjects of interest.
	
	\subsubsection{Trunk-Branch Ensemble CNN}
	
\begin{figure}[h]
		\begin{center}
			\includegraphics[width=1.0\textwidth]{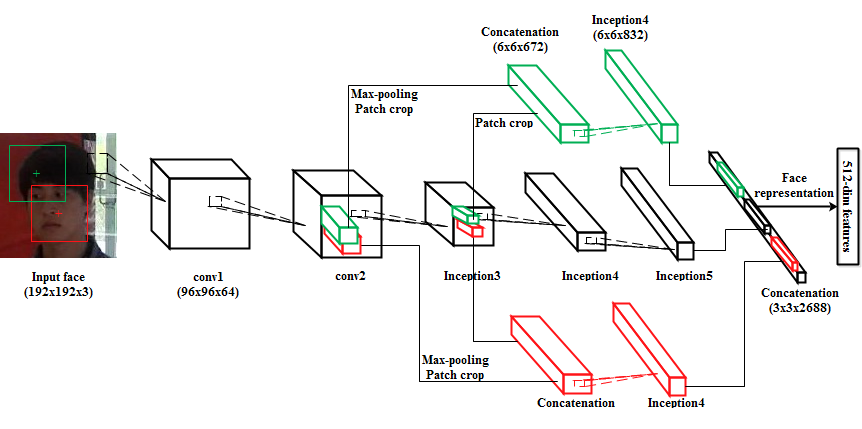}
		\end{center}
		\caption{The architecture of TBE-CNN \cite{Ding2016}.}
		\label{fig:TBE_CNN}
	\end{figure}

	An improved triplet-loss function has been introduced in \cite{Ding2016} to promote the robustness of face representations. To that end, a trunk-branch ensemble CNN (TBE-CNN) model has been proposed to extract complementary features from holistic face images, as well as, face patches around facial landmarks through trunk and branch networks, respectively. To emulate real-world video data, artificially blur training data are synthesized from still images by applying artificial out-of-focus and motion blur to learn blur-insensitive face representations. The architecture of TBE-CNN is shown in Figure \ref{fig:TBE_CNN}.
	
	As shown in Figure \ref{fig:TBE_CNN}, TBE-CNN contains one trunk network along with several branch networks, where the trunk and branch networks share some layers in order to embed global and local information. This sharing strategy may lead to reduce the computational cost and also efficient convergence. The output feature maps of these networks are concatenated to feed into the fully-connected layer to generate final face representations.
	
\begin{figure}[h]
		\begin{center}
			\includegraphics[width=0.97\textwidth]{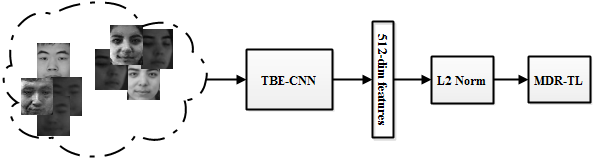}
		\end{center}
		\caption{Training pipeline of the TBE-CNN \cite{Ding2016}.}
		\label{fig:TBE_Training}
	\end{figure}

	During training as illustrated in Figure \ref{fig:TBE_Training}, TBE-CNN is given still images and simulated video frames, where the network aims to classify each still image and its corresponding artificially blurred face image correctly into the same class. The training process is performed using a stage-wise strategy, where the trunk network and each of the branch networks are trained separately with fixed parameters.
	
\begin{figure}[h]
		\begin{center}
			\includegraphics[width=0.67\textwidth]{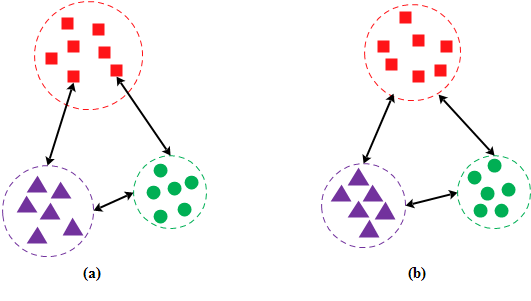}
		\end{center}
		\caption{The mean distance regularized triplet-loss. (a) Training triplet with non-uniform inter- and intra-class distance distributions, and (b) triplets with uniform inter- and intra-class distance distributions using MDR-TL regularization \cite{Ding2016}.}
		\label{fig:TBE_MDR}
	\end{figure}

	To improve the discriminative power of face representations, mean distance regularized triplet-loss (MDR-TL) function is considered to fine-tune the entire network. Compared to the original triplet-loss function proposed in \cite{Schroff_2015_CVPR}, MDR-TL regularizes the triplet-loss to provide uniform distributions for both inter- and intra-class distances. Figure \ref{fig:TBE_MDR} represents the principle of MDR-TL.
	
	As demonstrated in Figure \ref{fig:TBE_MDR}(a), it is difficult to appropriately discriminate between matching and non-matching pairs of face images because the training samples have non-uniform inter- and intra-class distance distributions. To tackle this problem, the triplet loss is regularized using MDR-TL loss function by constraining the distances between mean representations of different subjects (Figure \ref{fig:TBE_MDR}(b)).

	\subsubsection{HaarNet}
	
	An ensemble of deep CNNs called HaarNet has been proposed in \cite{Parchami2017} to efficiently learn robust and discriminative face representations for video-based FR applications. Similar to TBE-CNN \cite{Ding2016}, HaarNet consists of a trunk network with three diverging branch networks that are specifically designed to embed facial features, pose, and other distinctive features. The trunk network effectively learns a holistic representation of the face, whereas the branches learn more local and asymmetrical features related to pose or special facial features by means of Haar-like features. Furthermore, to increase the discriminative capabilities of the HaarNet, a second-order statistic regularized triplet-loss function has been introduced to take advantage of the inter-class and intra-class variations existing in training data to learn more distinctive representations for subjects with similar faces. Finally, a fine-tuning stage has been performed to embed the correlation of facial ROIs stored during enrollment and improve recognition accuracy.
	
\begin{figure*}[htp!]
		\begin{center}
			\includegraphics[width=1.0\textwidth]{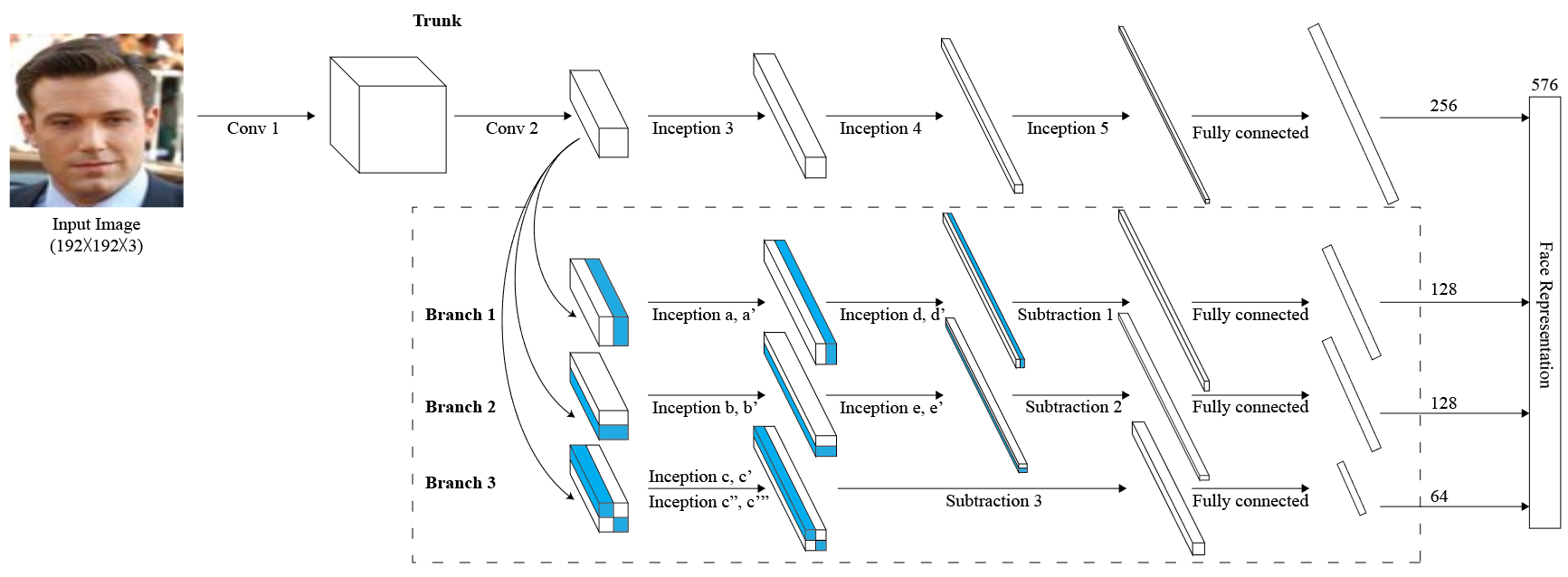}
		\end{center}
		\caption{HaarNet architecture for the trunk and three branches \cite{Parchami2017}. (Max pooling layers after each inception and convolution layer are not shown for clarity).}
		\label{fig:Network}
	\end{figure*}

	The overall architecture of the HaarNet is presented in Figure \ref{fig:Network}. It is composed of a global trunk network along with three branch networks that can effectively learn a representation that is robust to changing capture conditions.
	
	As shown in Fig.~\ref{fig:Network}, the trunk is employed to learn the global appearance face representation, whereas three branches diverged from the trunk are designed to learn asymmetrical and more locally distinctive representations. For the trunk network, the configuration of GoogLeNet \cite{Szegedy_2015_CVPR} is employed with 18 layers.
	
\begin{figure}[h]
		\centering
		\includegraphics[width=0.37\textwidth]{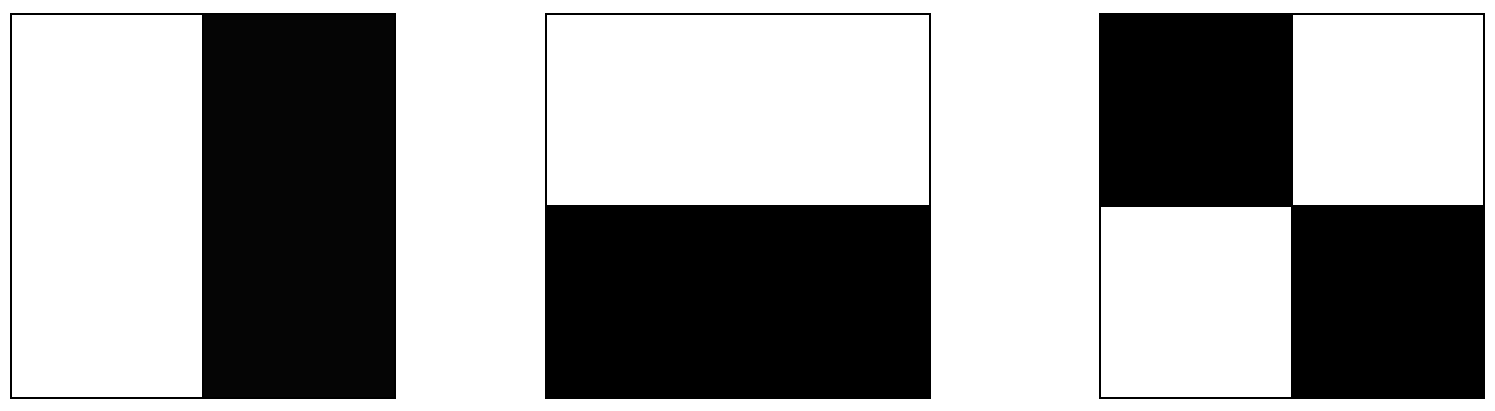}
		\caption{Haar-like features used in branch networks \cite{Parchami2017}.}
		\label{fig:Haar}
	\end{figure}

	In contrast with \cite{Ding2016}, instead of training each branch on different face landmarks, Haarnet utilizes three branch networks in order to compute one of the Haar-like features, respectively as illustrated in Fig. \ref{fig:Haar}. Haar features have been exploited to extract distinctive features from faces based on the symmetrical nature of facial components, and on contrast of intensity between adjacent components. In general, these features are calculated by subtracting sum of all pixels in the black areas from the sum of all pixels in the white areas. To avoid information loss, the Haar-like features are calculated by matrix summation, where black matrices are negated. Thus, instead of generating only one value, each Haar-like feature returns a matrix.
	
	In the Haarnet architecture (see Fig. \ref{fig:Network}), the trunk network and its three branches share the first two convolutional layers. Then, the first and second branches split the output of Conv2 into two sub-branches, and also apply two inception layers to each sub-branch. Subsequently, the two sub-branches are merged by a subtraction layer to obtain a Haar-like representation for each corresponding branch. Meanwhile, the third branch divides the output of Conv2 into four sub-branches and one inception layer is applied to each of the sub-branches. Eventually, a subtraction layer is exploited to combine those for sub-branches and feed to the fully connected layer. The final representation of the face is obtained by concatenating the output of the trunk and all three Haar-like features.
	
\begin{figure}[h]
		\centering
		\includegraphics[width=0.97\textwidth]{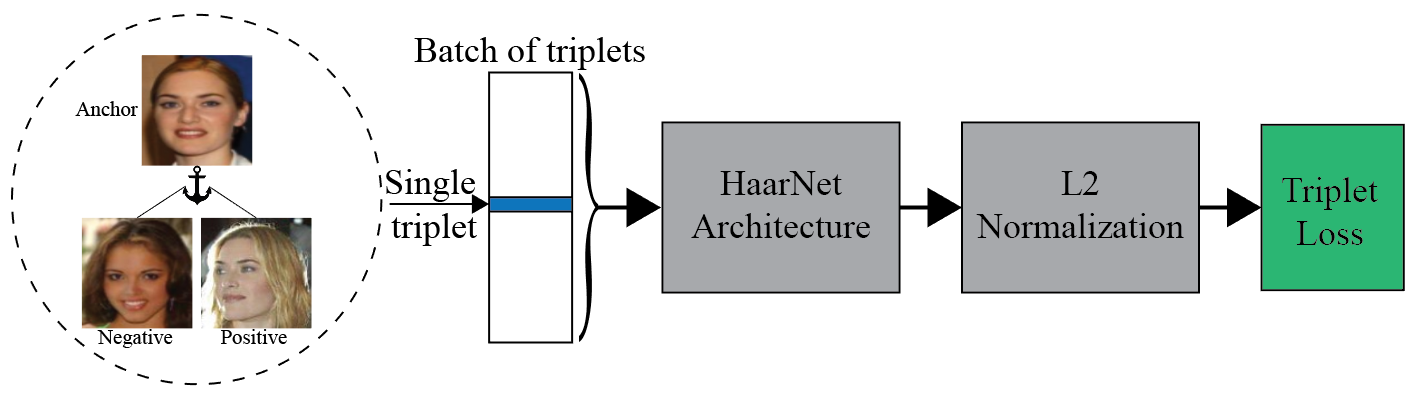}
		\caption{Processing of triplets to compute the loss function. The network inputs a batch of triplets to the HaarNet architecture followed by an $L2$ Normalization \cite{Parchami2017}.}
		\label{fig:Structure}
	\end{figure}

	Fig.~\ref{fig:Structure} illustrates the training process of the HaarNet using a triplet-loss concept, where a batch of triplets composed of <anchor, positive, negative> is input to the architecture is translated to a face representation.
	
	As shown in Fig. \ref{fig:Structure}, output of the HaarNet is then $L2$ normalized prior to feed into the triplet-loss function in order to represent faces on a unit hyper-sphere. Let’s denote the $L2$ normalized representation of a facial ROI $x$ as $f\left(x\right)\in\;{R^d}$ where $d$ is the dimension of the face representation.
	
	A multi-stage training approach is hereby considered to effectively optimize the parameters of the HaarNet. The first three stages are designed for initializing the parameters with a promising approximation prior to employ the triplet-loss function. Moreover, these three stages are beneficial to detect a set of hard triplets from the dataset in order to initiate the triplet-loss training. In the first stage, the trunk network is trained using a softmax loss, because the softmax function converges much faster than triplet-loss function. During the second stage, each branch is trained separately by fixing the shared parameters and by only optimizing the rest of the parameters. Similar to the first stage, a softmax loss function is used to train each of the branches. Then, the complete network is constructed by assembling the trunk and the three branch networks. The third stage of the training is indeed a fine-tuning stage for the complete network in order to optimize these four components simultaneously. In order to consider the inter- and intra-class variations, the network is trained for several epochs using the hard triplets detected during the previous stages.
	
\begin{figure}[h]
		\centering
		\includegraphics[width=0.67\textwidth]{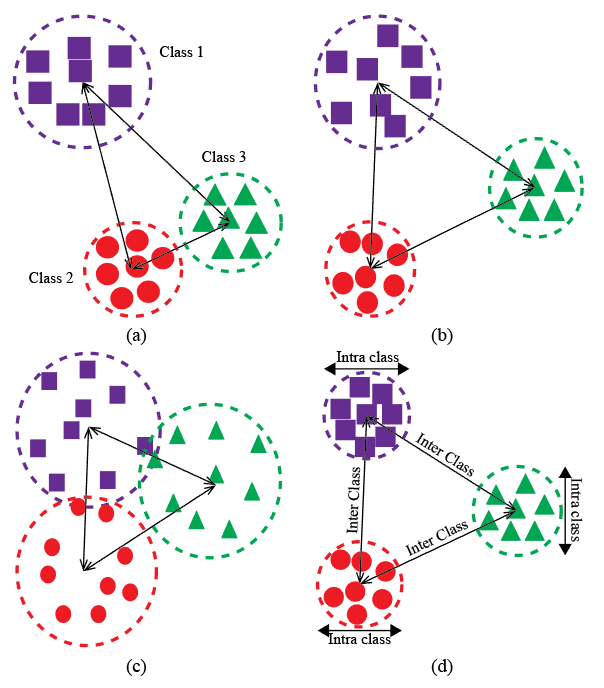}
		\caption{Illustration of the regularized triple loss principles based on the mean and standard deviation of three classes, assuming 2D representations of the ROIs \cite{Parchami2017}.}
		\label{fig:RegExample}
	\end{figure}

	As suggested in \cite{Ding2016}, adding mean distance regularization term to the triplet-loss function can promote distinctiveness of the face representations. Inspired from \cite{Ding2016}, the main idea of the second-order statistics regularization term is illustrated in Figure \ref{fig:RegExample} illustrates.
	
	In Fig~\ref{fig:RegExample} (a), triplet-loss function may suffer from nonuniform inter-class distances that leads to failure of using simple distance measures, such as Euclidean and cosine distances. In this regard (see Fig. \ref{fig:RegExample} (b)), a mean distance regularization term can be added to increase the separation of class representations. On the other hand, representations of some facial ROIs may be confused with representation of the adjacent facial ROIs in the feature space due to high intra-class variations. Fig. \ref{fig:RegExample} (c) shows such a configuration, where the mean representation of the classes are distant from each other but the standard deviations of classes are very high, leading to overlap among class representations. To address this issue, a new term in the loss function is introduced to examine the intra-class distribution of the training samples.
	
	The triplet constraint can be expressed as a function of the representation of anchor, positive and negative samples as follows \cite{Schroff_2015_CVPR}:
	
	\begin{equation}
	\left\| {f\left( {x_i^a} \right) - f\left( {x_i^p} \right)} \right\|_2^2 + a < \left\| {f\left( {x_i^a} \right) - f\left( {x_i^n} \right)} \right\|_2^2\
	\label{eq:1}
	\end{equation}
	\hspace*{-0.17cm} where $f\left( {x_i^a} \right)$, $f\left( {x_i^p} \right)$, and $f\left( {x_i^n} \right)$ are the face representations of the anchor, positive, and negative, respectively. All the triplets sampled from the training set should satisfy the constraint. Thus, during training, HaarNet minimizes of the loss function:
	
	\begin{equation}
	L_{HaarNet} = \;{\delta _1}{L_{triplet}} + \;{\delta _2}{L_{mean}} + \;{\delta _3}{L_{std}}
	\label{eq:2}
	\end{equation}
	\hspace*{-0.17cm} where ${\delta _i}$ denotes the weight for each term in the loss function. Furthermore, ${L_{triplet}}$ can be defined based on (\ref{eq:1}) as follows:
	
	\begin{equation}
	L_{triplet}=\frac{1}{{2N}} \mathop \sum \limits_{i = 1}^N {\left[ {\left\| {f\left( {x_i^a} \right) - f\left( {x_i^p} \right)} \right\|_2^2 - \left\| {f\left( {x_i^a} \right)-f\left( {x_i^n} \right)} \right\|_2^2 +\alpha} \right]_ + }
	\end{equation}
	
	Similar to \cite{Ding2016}, assuming that the mean distance constraint is $\beta  < \left\| {{{\hat \mu }_c} - \hat \mu _c^n} \right\|_2^2$, ${L_{mean}}$ is defined as:
	
	\begin{equation}
	{L_{mean}} = \;\frac{1}{{2P}}\;\mathop \sum \limits_{c = 1}^C {\rm{max}}\left( {0,\;\beta  - \;\left\| {{{\hat \mu }_c} - \hat \mu _c^n} \right\|_2^2} \right)
	\end{equation}
	
	In addition, the standard deviation constraint is defined to be ${\sigma _c} > \;\gamma $, where ${\sigma _c}$ is the standard deviation of the class $c$. Therefore, ${L_{std}}$ can be computed as follows:
	
	\begin{equation}
	{L_{std}} = \;\frac{1}{M}\;\mathop \sum \limits_{c = 1}^C {\rm{max}}\left( {0,\;\gamma  - \;{\sigma _c}} \right)
	\end{equation}
	\hspace*{-0.17cm} where $N$, $P$, and $M$ are the number of samples that violate the triplet, mean distance, and standard deviation constraints, respectively. Likewise, $C$ is the number of subjects in the current batch and $\alpha $, $\beta $,and $\gamma $ are margins for triplet, mean distance, and standard deviation constraints, respectively. The loss function (\ref{eq:2}) can be optimized using the regular stochastic gradient descent with momentum similar to \cite{Ding2016}. The gradient of loss w.r.t. the facial ROI representation of $i$th image for subject $c$ (denoted as $f\left( {{x_{ci}}} \right)$) is derived as follows:
	
	\begin{equation}
	\frac{{\partial {L_{std}}}}{{\partial f\left( {{x_{ci}}} \right)}} =  - \;\frac{1}{M}\;\mathop \sum \limits_{c = 1}^C {\omega _c}\;\frac{{\partial {\sigma _c}}}{{\partial f\left( {{x_{ci}}} \right)}}\;
	\end{equation}
	\hspace*{-0.17cm} where ${\omega _c}$ equals to 1 if the standard deviation constraint is violated, and equals to 0 otherwise. Moreover, the derivative of ${L_{std}}$ can be computed by applying the chain rule as follows:
	
	\begin{equation}
	\begin{split}
	\frac{{\partial {\sigma _c}}}{{\partial f\left( {{x_{ci}}} \right)}} = \;\frac{{\partial \sqrt {\frac{1}{{{N_c}}}\mathop \sum \nolimits_{j = 1}^{{N_c}} \left\| {f\left( {{x_{cj}}} \right) - \;{\mu _c}} \right\|_2^2} }}{{\partial f\left( {{x_{ci}}} \right)}} = \\
	\;\frac{{\left[ {\mathop \sum \nolimits_{j = 1}^{{N_c}} \frac{1}{{{N_c}}}{{\left\| {{\mu _c} - f\left( {{x_{cj}}} \right)} \right\|}_2}} \right] - {{\left\| {{\mu _c} - f\left( {{x_{ci}}} \right)} \right\|}_2}}}{{2\;\sqrt {\frac{1}{{{N_c}}}\mathop \sum \nolimits_{j = 1}^{{N_c}} \left\| {f\left( {{x_{cj}}} \right) - \;{\mu _c}} \right\|_2^2} }}\
	\end{split}
	\end{equation}
	
	As shown in Fig. \ref{fig:RegExample} (d), the discriminating power of the face representations can be improved by setting margins such that $\gamma < \beta $. This ensures a high inter-class and a low intra-class variations to increase the overall classification accuracy.
	
	\subsection{Deep CNNs Using Autoencoder}
	
	An efficient Canonical Face Representation CNN (CFR-CNN) has been proposed in \cite{Parchami2017CFR} for accurate still-to-video FR from a SSPP, where still and video ROIs are captured under various conditions. The CFR-CNN is based on a supervised autoencoder that can represent the divergence between the source (still ROI) and target (video ROI) domains encountered in still-to-video FR scenario. The autoencoder network is trained using a weighted pixel-wise loss function that is specialized for SSPP problems, and allows to reconstruct canonical ROIs (frontal and less blurred faces) for matching that correspond to the conditions of reference still ROIs. In addition, it can generate discriminative face embeddings that are similar for the same individuals, and robust to variations typically observed in unconstrained real-world video scenes. A fully-connected classification network is also trained to perform face matching using the face embeddings extracted from the deep autoencoder, and accurately determine whether the pairs of still and video ROIs correspond to the same individual.
	
\begin{figure}[h]
		\begin{center}
			\includegraphics[width=0.97\textwidth]{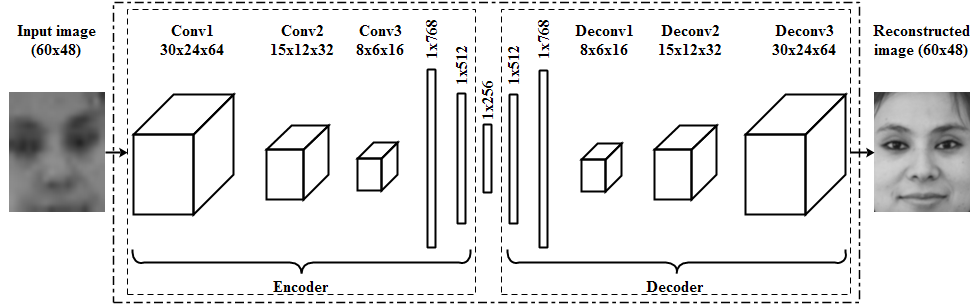}
		\end{center}
		\caption{Block diagram of the autoencoder network in the CFR-CNN \cite{Parchami2017CFR}.}
		\label{fig:AutoNetwork}
	\end{figure}

	Autoencoder CNNs are typically utilized to normalize variations in face capture conditions from probe video ROIs to those in still reference ROIs. The architecture of the autoencoder is shown in Figure~\ref{fig:AutoNetwork}, where the input image is a probe video ROI captured using a surveillance camera, while the output is a reconstructed image. This network consists of (1) three convolutional layers each followed by a max-pooling layer to extract robust convolutional maps, and then (2) a two-layer fully-connected network that generates a 256-dimensional face embedding. The decoder reverses these operations by applying a fully-connected layer to generate the original vector and three deconvolutional layers, each one followed by un-pooling layers designed for generating the final reconstruction of the face.
	
\begin{figure}[h]
		\begin{center}
			\includegraphics[width=0.17\textwidth]{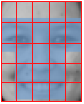}
		\end{center}
		\caption{T-shaped weight mask used for the loss function of CFR-CNN \cite{Parchami2017CFR}.}
		\label{fig:TShape}
	\end{figure}

	A development set (assumed to be collected from unknown individuals captured from the operational domain) is employed for training of the deep autoencoder network. A batch of video ROIs are fed into the network, where still ROIs of the corresponding persons are used for facial reconstructions. Using higher-quality still images that are captured during enrollment under controlled conditions as target faces, the autoencoder network simultaneously learns invariant face embeddings to normalize the input video ROIs. The parameters of this autoencoder network are optimized by employing a weighted Mean Squared Error (MSE) criterion, where a T-shaped region (illustrated in Figure~\ref{fig:TShape}) is considered to assign a higher significance to discriminative facial components like eyes, nose and mouth. This loss function of is formulate as:
	\begin{equation}
	\begin{aligned}
	{L_{CFR - CNN}} = \mathop \sum \limits_{ i \in rows } \mathop \sum \limits_{ j \in cols } {\tau _{i,j}}\left\| {{X^2} - {{\hat X}^2}} \right\| \\
	\tau_{i,j} = \left\{
	\begin{matrix*}[l]
	\alpha & \text{if (i,j) belongs to T}\\ 
	\beta & \text{if (i,j) otherwise}\\
	\end{matrix*}\right.
	\end{aligned}
	\end{equation}
	\hspace*{-0.17cm} where $rows\times cols$ is the size of ROIs, $X$ is the target still ROI and $\hat{X}$ is the reconstructed ROI. The weight $\alpha$ is considered for the T region, while the weight $\beta$ is considered for pixels outside the T region.
	
\begin{figure}[h!]
		\begin{center}
			\includegraphics[width=0.67\textwidth]{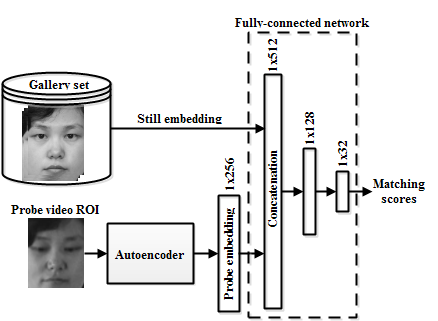}
		\end{center}
		\caption{Block diagram of the classification network in the CFR-CNN \cite{Parchami2017CFR}.}
		\label{fig:classification}
	\end{figure}

	A fully-connected network is then integrated with the deep convolutional autoencoder, and the output of the intermediate layer is then considered as a face representation that is invariant to the different nuisance factors commonly encountered in unconstrained surveillance environments. Finally, face matching is performed using a fully-connected classification network as shown in Figure \ref{fig:classification}. This network is implemented to match the face representations of still and video ROIs.
	
	The fully-connected classification network is trained using a regular pairwise-matching scheme, where the face embeddings of the reference still and probe video ROIs are fed into the classification network. The network can thereby learn to classify each pair of still and video ROIs as either matching or non-matching.
	
	\section{Performance Evaluation}
	\label{sec:Experiment}
	
\begin{figure}[h]
		\begin{center}
			\includegraphics[width=0.97\textwidth]{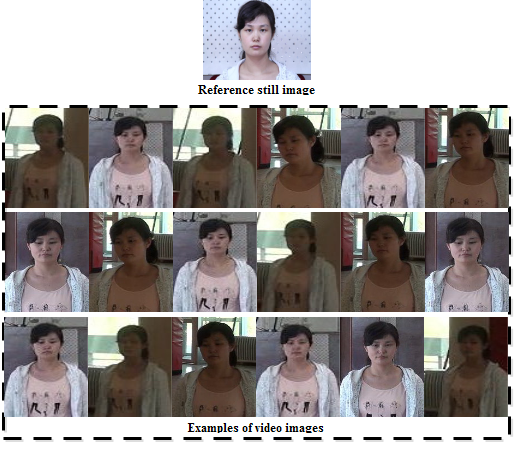}
		\end{center}
		\caption{An example of high-quality reference still image and random low-quality video images of the corresponding individual captured by the still camera and three camcorders in the COX Face DB \cite{Huang2015}.}
		\label{fig:CoxFace}
	\end{figure}

	The performance of the aforementioned video-based FR systems is evaluated using Cox Face DB~\cite{Huang2015}. This dataset was specifically collected for video surveillance applications, where it is composed of high-quality still faces captured with still cameras under controlled conditions and low-quality video faces captured with different off-the-shelf camcorders under uncontrolled conditions. Videos are recorded per subject when they are walking through a designed-S curve containing changes in pose, illumination, scale and blur. An example of still and videos of one subject is shown in Figure \ref{fig:CoxFace}.
	
	The systems are evaluated according to experimental protocol suggested in \cite{Huang2015}, where each probe video ROI is compared against the reference still ROIs, and rank-1 recognition is reported as the FR accuracy. Meanwhile, since video-based FR systems are often required to perform real-time processing in surveillance applications, the computational complexity of such systems should be also taken into consideration. In this regards, the complexity can be determined in terms of the number of operations (to match a video probe ROI to a reference still ROI), the number of network parameters and layers \cite{Canziani2016}.
	
	In order to confirm the viability of the CNN-based video FR systems for real-time surveillance applications, Table \ref{tbl:Results} presents the accuracy and compares their computational complexity.
	
	\begin{table}[h]
		\centering
		\caption{Rank-1 recognition and computational complexity of video-based FR systems over videos of Cox Face DB.}
		\label{tbl:Results}
		\begin{adjustbox}{width=0.87\textwidth}
			\begin{tabular}{|l||c||c|c|c|}
				\hline
				\multirow{2}{*}{\textbf{FR system}} & \textbf{Rank-1} & \multicolumn{3}{c|}{\textbf{Computational complexity}}                      \\ \cline{3-5} 
				& \textbf{recognition}    & \textbf{\# operations} & \textbf{\# parameters} & \textbf{\# layers} \\ \hline \hline
				\textbf{CCM-CNN} \cite{Parchami2017CCM}                     & 89.53\begin{math}\pm\end{math}0.9           &      33.3M                         & 2.4M          &  30               \\ \hline
				\textbf{TBE-CNN} \cite{Ding2016}                     & 90.61\begin{math}\pm\end{math}0.6                  & 12.8B                               & 46.4M           &  144              \\ \hline
				\textbf{HaarNet} \cite{Parchami2017}                    & 91.40\begin{math}\pm\end{math}1.0           &             3.5B                  & 13.1M        & 56                 \\ \hline
				\textbf{CFR-CNN} \cite{Parchami2017CFR}                    & 87.29\begin{math}\pm\end{math}0.9           &      3.75M                         & 1.2M          &  7               \\ \hline
			\end{tabular}
		\end{adjustbox}
	\end{table}
	
	It can be seen in Table \ref{tbl:Results} that the TBE-CNN and HaarNet provide the highest level of accuracy, while they are very complex. Although the CCM-CNN and CFR-CNN cannot outperform these deep architectures, but they can achieve satisfactory results with significantly lower computational complexity. Moreover, the number of network parameters and layers are key factors in designing deep CNN that can greatly affect the convergence and training time. Considering these criteria, the proposed CCM-CNN and CFR-CNN have the lowest design complexity, and subsequently the shortest convergence time.
	
	\section{Conclusion and Future Directions}
	\label{sec:Future}
	
	In this chapter, the most recent deep learning architectures proposed for robust face recognition in video surveillance were thoroughly investigated. To overcome the existing challenges in real-world surveillance unconstrained environments, the single training reference sample and domain adaptation problems have been taken into account during the system design. On the other hands, computational complexity is also a key issue to provide an efficient solution for real-time video-based FR systems. In particular, this chapter reviewed deep learning architectures proposed based on triplet-loss function and autoencoder CNNs.
	
	Triplet-based loss optimization method allows to learn complex and non-linear facial representations that provide robustness across inter- and intra-class variations. CCM-CNN proposes a cost-effective solution that is specialized for still-to-video FR from a single reference still by simulating weighted CCM. TBE-CNN and HaarNet can extract robust representations of the holistic face image and facial components through an ensemble of CNNs containing one trunk and several branch networks. In addition, to compensate the limited robustness of facial model in the case of single reference still, they were fine-tuned using synthetically-generated faces from still ROIs of non-target individuals. In contrast, CFR-CNN employed a supervised autoencoder CNN to generate canonical face representations from low-quality video ROIs. It can therefore reconstruct frontal faces that correspond to capture conditions of reference still ROIs and generate discriminant face representations. Experimental results obtained with the COX Face DB indicated that TBE-CNN and HaarNet can achieve higher level of accuracy with heavy computational complexity, while CCM-CNN and CFR-CNN can provide convincing performance with significantly lower computational costs.
	
	Since the use of deep learning is increasingly growing, one of the future direction is to integrate conventional methods with deep learning methods in order to incorporate statistical and geometrical properties of faces into the deep features. In addition, future research can focus on utilizing temporal information, where facial ROIs can be tracked over frames to accumulate the predictions over time. Thus, the combination of face detection, tracking, and classification in a unified deep learning-based network will lead to a robust spatio-temporal recognition suitable for real-world video surveillance applications. Thus, 3D CNNs and recurrent neural networks such as long short-term memory can be exploited to consider convolutions through the time, due to capturing temporal information among successive video frames.
	
	\begin{acknowledgement}
		This work was supported by the Fonds de Recherche du Qu\'ebec - Nature et Technologies and MITACS.
	\end{acknowledgement}
	
	\bibliographystyle{spmpsci}
	\bibliography{Deepbib}
	
\end{document}